\documentclass{article}

\PassOptionsToPackage{numbers, sort&compress}{natbib}

\usepackage[preprint]{neurips_2026}

\usepackage[utf8]{inputenc} 
\usepackage[T1]{fontenc}    
\usepackage{microtype}      

\usepackage{amsmath}
\usepackage{amssymb}

\usepackage{booktabs}
\usepackage{multirow}
\usepackage{array}
\usepackage{tabularx}

\usepackage{algorithm}
\usepackage{algpseudocode}

\usepackage{enumitem}

\usepackage{float}

\usepackage{xspace}

\usepackage{graphicx}
\graphicspath{{figures/}}
\usepackage{xcolor}
\definecolor{LinkBlue}{HTML}{1A73E8}
\definecolor{figtextcolor}{HTML}{333333}

\usepackage{xurl}
\usepackage[colorlinks=true, allcolors=LinkBlue]{hyperref}

\def\NickName{Self Flow Forcing}
\def\NickNameShort{SF\textsuperscript{2}}
\newcommand{\websiteurl}{https://sotamak1r.github.io/SF2-web/}

\definecolor{mycolor1}{RGB}{72, 146, 232}
\definecolor{mycolor2}{RGB}{142, 120, 204}
\definecolor{mycolor3}{RGB}{197, 103, 124}

\title{Beyond Distribution Matching: Self-Supervised Representation Forcing for Few-Step Video Generation}

\author{%
  Junyi Chen$^{1,}$\thanks{Work done during an internship at Hunyuan, Tencent.}\quad
  Haiyu Zhang$^{2}$ \quad
  Zhoujie Fu$^{3}$ \quad
  Tengfei Wang$^{2,}$\thanks{Corresponding author.} \quad
  Chunchao Guo$^{2}$ \\[0.6em]
  $^{1}$Shanghai Jiao Tong University \quad
  $^{2}$Hunyuan, Tencent \quad
  $^{3}$Nanyang Technological University
}

\begin{document}

\maketitle

\begin{abstract}

Few-step distillation is essential for deploying modern video diffusion models. 
Among existing approaches, Distribution Matching Distillation (DMD) has emerged as a highly promising paradigm by achieving strong output-space fidelity without adversarial training, while remaining flexible across various student architectures.
We observe, however, that DMD fundamentally relies on an output-space objective, lacking explicit regularization on the model's internal representations. 
As the distillation step count aggressively shrinks, this output-only supervision becomes insufficient, 
causing the distilled student to degrade in both fine-grained spatial structures and coherent temporal dynamics.
To address this limitation, we introduce \NickName{} (\NickNameShort), which incorporates a predictive representation loss into the DMD framework without altering its output objective. 
By providing the student and an Exponential Moving Average (EMA) teacher with heterogeneous noise levels, we establish an information asymmetry. 
This formulation forces the student to predict the teacher's internal representations from a more corrupted input, effectively integrating self-supervised representation alignment into distribution matching.
Extensive experiments demonstrate that \NickNameShort{} effectively mitigates temporal flickering and consistently improves both spatial fidelity and motion quality. 
Furthermore, the proposed mechanism is architecture-agnostic—seamlessly supporting both bidirectional and causal models—and requires no external feature extractors, introducing minimal computational overhead.

Project page: \url{https://sotamak1r.github.io/SF2-web/} 

Code: \url{https://github.com/SOTAMak1r/SF2/} 

\end{abstract}


\section{Introduction}
\label{sec:introduction}

Video diffusion models~\cite{vdm,svd,sora,makeavideo,vino} have recently delivered impressive advances in synthesizing high-fidelity, dynamic visual content. Yet, their iterative generation process typically requires tens to hundreds of denoising steps, posing a significant obstacle to real-world deployment. 
To bridge this gap, various few-step distillation paradigms have been explored, including Consistency Models~\cite{cm, lcm}, Trajectory Straightening~\cite{reflow, instaflow}, and Adversarial Distillation~\cite{add}. 
Among these, Distribution Matching Distillation (DMD)~\cite{dmd,dmd2} has emerged as a highly promising approach by achieving strong output-space fidelity,
while seamlessly supporting various student architectures.

However, when DMD is applied to the video domain, a characteristic degradation emerges at very low step counts. 
Specifically, the few-step distilled student suffers along two critical axes of video quality: fine-grained spatial structures (e.g., text rendering, complex geometry, and small object details) become blurred or inconsistent, and motion exhibits pronounced temporal flickering and physically unnatural dynamics. 
We observe this dual degradation in both our bidirectional and causal baselines, suggesting that it stems from the underlying distillation formulation itself rather than specific architectural choices.

We attribute this degradation to DMD's exclusive reliance on output-space supervision. 
DMD trains the student by matching its marginal samples to the teacher's at the output level, but it lacks explicit regularization to enforce the alignment of internal representations. 
In standard multi-step diffusion, this output-level proxy is sufficient, as the student can gradually refine its predictions and absorb fine-grained structures through iterative denoising. 
As the step budget aggressively shrinks, however, this proxy becomes inadequate. 
The student lacks the iterative capacity to recover rich, structured internal features, leading to the observed deterioration in both spatial fidelity and temporal coherence.

A complementary perspective from self-supervised representation learning offers a natural solution to this structural degradation.
Classic predictive frameworks, from masked autoencoding~\cite{mae_st,videomae} to V-JEPA~\cite{vjepa2}, have established that representation-level prediction enforces robust spatiotemporal understanding.
Recent methods~\cite{repa,selfflow,sra} 
have successfully adapted this principle for generative models, proving that internal feature regularization directly translates to higher generation fidelity.
DMD, by contrast, is entirely devoid of such explicit representation-level guidance, optimizing only the marginal output distribution.

Motivated by this observation, we introduce \NickNameShort, a novel paradigm that injects a predictive representation objective directly into the DMD framework without altering its output loss. 
The core mechanism is to establish an information asymmetry between the student and an Exponential Moving Average (EMA) teacher. 
By feeding the two networks with heterogeneous noise levels, we present the student with a more corrupted view of the underlying sequence and force it to predict the teacher's robust internal representations. 
This formulation explicitly integrates self-supervised representation alignment into distribution matching. 
Importantly, our mechanism requires no external feature extractors—the teacher is a self-bootstrapped EMA of the student—and adds only one extra forward pass per iteration, remaining agnostic to the specific architectural form of the generator.

We validate \NickNameShort{} across two representative video distillation settings: a bidirectional text-to-video student and a causal autoregressive student, building upon the standard ODE-initialization protocols~\cite{causalforcing}. 
In both scenarios, \NickNameShort{} effectively mitigates temporal flickering and consistently improves both spatial fidelity and motion quality with minimal computational overhead.

In summary, our main contributions are:

\begin{itemize}[left=0pt]
\item We identify a critical limitation of applying standard DMD to video generation: its exclusive reliance on output-space supervision lacks explicit regularization for the student's internal representations. In the few-step regime, this results in the widely observed spatial artifacts and temporal flickering.

\item We propose \NickNameShort, which incorporates a predictive representation objective into the DMD framework. By utilizing heterogeneous noise levels and an EMA teacher to create an information asymmetry, we force the student to predict the teacher's internal representations from a more corrupted view, effectively integrating representation-level alignment with distribution matching.

\item Extensive experiments on bidirectional and causal video distillation settings demonstrate that \NickNameShort{} consistently enhances spatial fidelity and temporal coherence, all while requiring no external feature extractors and introducing minimal computational overhead.

\end{itemize}

\section{Related Work}
\label{sec:related_work}

\NickName\space sits at the intersection of two rapidly evolving domains: the acceleration of video diffusion models through step distillation, and the enhancement of generative paradigms via self-supervised representation learning. We briefly review both areas and position our contribution.

\subsection{Efficient Video Distillation}

Step distillation aims to compress the iterative sampling process of diffusion models into a few steps. 
In the image domain, paradigms such as Progressive Distillation~\cite{progressiedistillation}, Consistency Models~\cite{cm,lcm}, Adversarial Diffusion Distillation~\cite{add}, flow-based trajectory straightening~\cite{reflow,instaflow}, and Distribution Matching Distillation (DMD)~\cite{dmd,dmd2} have established the feasibility of few-step generation.
These acceleration frameworks have subsequently been extended to the video domain. 
This includes bidirectional text-to-video adaptations such as VideoLCM~\cite{videolcm}, T2V-Turbo~\cite{t2vturbo}, and PCM~\cite{pcm}.
In parallel, causal autoregressive generation frameworks have been accelerated through CausVid~\cite{causvid}, Self-Forcing~\cite{selfforcing}, and Causal-Forcing~\cite{causalforcing}.
However, a persistent commonality across these methods is their exclusive reliance on marginal output distribution supervision, leaving the student's internal representations completely unconstrained. 
We identify this output-only supervision as the fundamental structural cause of the spatial degradation and temporal flickering widely observed in few-step distilled video models. 
\NickName\space addresses this limitation by introducing a representation-level objective that operates orthogonally to the existing output objective.

\subsection{Self-Supervised Representation Learning in Generative Models}

A growing body of work demonstrates that integrating representation-level priors into generative training yields cleaner internal features and superior generation quality. 
This integration typically follows two distinct routes. 
The first route injects semantic priors directly into the latent space: methods such as VA-VAE~\cite{vavae}, SemanticGen~\cite{semanticgen} and RAE~\cite{rae} align or replace the standard VAE latents with rich semantic features from frozen external perception encoders~\cite{dinov2,siglip2,qwen25vl}. 
The second route operates within the generative backbone itself: REPA~\cite{repa, repae} aligns the intermediate layers of a diffusion transformer with these frozen external encoders during training. 
To remove the dependency on external extractors, SRA~\cite{sra} provides representation guidance via self-distillation, while Self-Flow~\cite{selfflow} integrates a predictive objective by feeding heterogeneous noise levels across tokens, creating an information asymmetry the network must inherently resolve.

This predictive, asymmetric paradigm is particularly crucial for the video domain. 
As established by architectures like V-JEPA~\cite{vjepa2} and VideoMAE~\cite{videomae}, learning to predict the representations of masked (perturbed) observations serves as a powerful inductive bias for capturing complex spatiotemporal dynamics.
However, while the aforementioned representation strategies successfully enhance the training of base multi-step generative models
---which operate with large step budgets and extensive data
---their potential within the highly compressed regime of few-step model distillation remains entirely unexplored. 
\NickName\space bridges this gap. By porting the predictive representation-forcing principle into video distillation via asymmetric noise, we force the student to capture robust spatiotemporal dynamics, effectively repairing the degradation of DMD without relying on external encoders.

\section{Preliminaries}
\label{sec:preliminaries}

Modern video generators are dominated by diffusion and flow-matching models~\cite{ldm,flowmatching}, which produce high-quality samples at the cost of tens to hundreds of sequential evaluation steps. 
Distribution Matching Distillation (DMD) addresses this inference bottleneck by compressing such multi-step generators into few-step students via score matching in the output space. 
While this serves as the universal foundation for our distillation framework, evaluating few-step causal autoregressive models introduces an additional train--test mismatch, which is effectively resolved by Self-Forcing~\cite{selfforcing}. \NickNameShort{} builds upon these foundational concepts, which we briefly review below.

\paragraph{Diffusion models.}
Given clean data $x_0 \sim p_{\text{data}}$, a generative forward process interpolates between the data distribution and a tractable noise distribution~\cite{ddpm,sgm}:
\begin{equation}
    x_t \;=\; \alpha_t \, x_0 \,+\, \sigma_t \, \epsilon, \qquad \epsilon \sim \mathcal{N}(0,\mathbf{I}), \qquad t \in [0, T],
    \label{eq:forward}
\end{equation}
where the schedule $(\alpha_t, \sigma_t)$ transports $x_0$ at $t{=}0$ to nearly pure noise at $t{=}T$.
Flow matching~\cite{flowmatching}, the de facto choice for modern video generators, fits the same template under a linear schedule.
A neural network is trained to invert this process by predicting an interconvertible target—such as the noise $\epsilon$, the clean data $x_0$, or the velocity $v$—all of which fundamentally parameterize the score $\nabla_{x_t} \log p_t(x_t)$~\cite{sgm,edm}. 
At inference time, however, generation requires integrating this vector field over tens to hundreds of timesteps, creating a severe sampling bottleneck.

\paragraph{Distribution Matching Distillation.}
DMD~\cite{dmd, dmd2} addresses this bottleneck by training a few-step student generator $G_\theta$. It maps a noisy latent $x_t$ at timestep $t$ directly to a clean prediction $\hat{x}_0 = G_\theta(x_t, t)$, such that the student's output distribution $p_\theta$ matches the target distribution $p_{\text{real}}$ implied by the frozen multi-step teacher. 
To bypass intractable high-dimensional distribution matching, DMD minimizes the reverse KL divergence:
\begin{equation}
    \mathcal{L}_{\text{DMD}} \;=\; \mathbb{E}_t\!\left[ D_{\text{KL}}\!\left(p_\theta \,\Vert\, p_{\text{real}}\right) \right]
    \label{eq:dmd_kl}
\end{equation}
through score matching at intermediate noise levels. Two score networks are maintained alongside the student: a frozen real-score critic $s_{\text{real}}(x_t, t)$ inherited from the teacher, and a learnable fake-score critic $s_\psi(x_t, t)$ tracking the student's current output distribution and is updated via the standard denoising loss. 
The resulting DMD gradient applied to the student is:
\begin{equation}
    \nabla_\theta \mathcal{L}_{\text{DMD}} \;=\; \mathbb{E}_{x_t, t}\!\left[\, \omega(t)\, \bigl(s_\psi(x_t, t) - s_{\text{real}}(x_t, t)\bigr)\, \nabla_\theta G_\theta(x_t, t) \,\right],
    \label{eq:dmd_grad}
\end{equation}
where $\omega(t)$ is a timestep weighting. 
Intuitively, this gradient pushes the student's outputs away from regions where its own samples are more likely than the teacher's. 
Critically for our motivation, this learning signal is derived entirely from the marginal output space $\hat{x}_0$, leaving the student's intermediate representation layers unregularized.

\paragraph{Self-Forcing for causal video distillation.}
While standard DMD is sufficient for bidirectional video distillation, causal autoregressive video generators face an additional context-conditioning challenge. 
Standard teacher forcing conditions every generated chunk on clean ground-truth history, causing severe error accumulation during autoregressive inference. 
Diffusion forcing~\cite{diffusionforcing} relaxes this by assigning independent noise levels per chunk, yet the conditioning history still relies on real data rather than the model's own predictions. 
Self-Forcing~\cite{selfforcing} closes this train--test gap by rolling the student out chunk-by-chunk on its own previously generated context. By propagating information via a key-value (KV) cache, training and inference share the exact same conditioning distribution. 
We adopt this protocol as the foundation for our causal autoregressive experiments.


\section{\NickName{} (\NickNameShort)}
\label{sec:method}

\begin{figure}
    \centering
    \includegraphics[width=\linewidth]{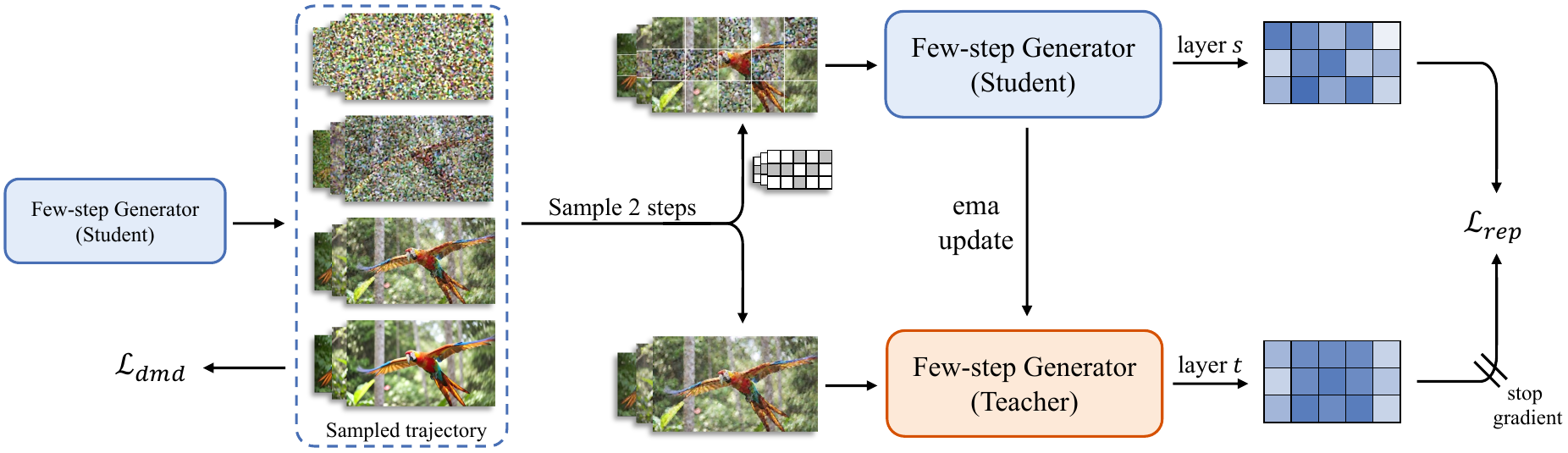}
    \vspace{-5mm}
    \caption{%
        \textbf{Overview of \NickNameShort{}.}
        From the student's denoising trajectory, the cleanest prediction drives the standard DMD loss $\mathcal{L}_{\text{DMD}}$, while two intermediate states $(x_t, x_s)$ with $t > s$ feed the \NickNameShort{} branch: a spatial mask mixes them into an asymmetric, highly corrupted input for the student, while the EMA teacher sees only the cleaner $x_s$. Hidden states from asymmetric layers are aligned via learnable projectors with a stop-gradient cosine loss $\mathcal{L}_{\text{rep}}$. The teacher is updated by EMA and requires no external pre-trained features.
    }
    \label{fig:pipeline}
\end{figure}


In this section, we present \NickNameShort{}, a framework for few-step video distillation. 
As identified in Sec.~\ref{sec:preliminaries}, a fundamental limitation of DMD is its exclusive reliance on output-level supervision, leaving the internal representations entirely unconstrained.
Under aggressive few-step compression, this unconstrained freedom becomes harmful: the student tends to converge to shortcut representations that merely minimize the marginal distribution loss, fundamentally lacking temporal and structural coherence.
\NickNameShort{} closes this gap by aligning the student's intermediate features with those of an exponential-moving-average (EMA) teacher using deliberately asymmetric inputs, while preserving the original DMD output objective. 
Concretely, we maintain three modules during training: the student $G_\theta$, an EMA teacher $G_{\theta'}$ with identical architecture but no gradients, and the standard DMD critics $s_\psi$ and $s_{\text{real}}$. 
The full training objective is formulated as:
\begin{equation}
    \mathcal{L}_{\text{total}} \;=\; \mathcal{L}_{\text{DMD}} + \lambda_{\text{rep}}\,\mathcal{L}_{\text{rep}},
    \label{eq:total_loss}
\end{equation}
where $\mathcal{L}_{\text{rep}}$ reuses the rollout that DMD already performs. 
Thus, the additional computational cost is dominated by a single teacher forward pass and remains independent of the number of denoising steps.

Following the pipeline illustrated in Fig.~\ref{fig:pipeline}, we structure the remainder of this section around three core components. 
We first detail the construction of asymmetric inputs from a shared trajectory (Sec.~\ref{sec:method_asymmetry}).
We then specify the representation loss $\mathcal{L}_{\text{rep}}$ (Sec.~\ref{sec:method_reploss}).
Finally, we detail the EMA teacher update mechanism and projector warmup that stabilize this self-bootstrapped alignment (Sec.~\ref{sec:method_ema}).

\subsection{Asymmetric Inputs from a Shared Trajectory}
\label{sec:method_asymmetry}

In standard generative training, uniform noise is typically applied across all spatial-temporal tokens. 
The resulting denoising objective can often be minimized by exploiting low-level, local correlations, providing little incentive for the model to capture deep semantic structures. 
To break this local dependency, we draw inspiration from Self-Flow~\cite{selfflow} and introduce  information asymmetry. 
The core idea is to apply heterogeneous noise levels across different tokens, forcing the student to proactively infer heavily corrupted regions using the global context provided by cleaner tokens.

However, introducing this asymmetry naively---such as completely masking out a subset of tokens or assigning independent, randomized noise levels to every token---severely disrupts the underlying generative dynamics and creates a significant train--inference mismatch. 
To safely establish this asymmetry without compromising the generative pipeline, we adopt a dual-timestep formulation. 
Starting from a shared denoising trajectory, the student receives a token-level mixture of a noisier state and a cleaner state, whereas the teacher observes exclusively the cleaner state. Consequently, the student must predict the teacher's representations from a strictly more corrupted vantage point, circumventing trivial self-reconstruction while seamlessly integrating representation learning into distribution matching.

\paragraph{Two trajectory states.}
The DMD generator loss already unrolls the student over $K$ denoising steps $t_K > \dots > t_1$,
\begin{equation}
    \hat{x}_0^{(k)} = G_\theta(x_{t_k}, t_k), \qquad
    x_{t_{k-1}} = \alpha_{t_{k-1}}\, \hat{x}_0^{(k)} + \sigma_{t_{k-1}}\, \epsilon^{(k)},
    \label{eq:rollout}
\end{equation}
producing a denoising trajectory $\{(x_{t_k}, t_k)\}_{k=1}^{K}$ of progressively cleaner latents. 
From this trajectory, we sample two distinct steps $(t, s)$ with $t > s$, giving a noisier latent $x_t$ and a cleaner latent $x_s$. 
Because DMD is an inherently data-free distillation paradigm that relies on self-generated samples, representation alignment requires us to first synthesize valid latent states. 
By repurposing the trajectory already unrolled for the standard DMD objective, we efficiently obtain these intermediate states without incurring the heavy computational cost of an independent generative rollout.

\paragraph{Per-token spatial masking.}
Let the latent video feature have $F$ frames, each tokenized into an $H_p \times W_p$ patch grid. 
We sample a binary mask $M \in \{0,1\}^{B \times F \times H_p \times W_p}$ where each entry drawn as $\mathrm{Bern}(\rho)$, $\rho \sim \mathcal{U}(0, \rho_{\max})$. 
The student input is formulated as a per-token mixture of the two trajectory states, whereas the teacher observes exclusively the cleaner state:
\begin{equation}
    \tilde{x}^{\text{stu}} = M \odot x_s + (1 - M) \odot x_t, \qquad
    \tilde{x}^{\text{tea}} = x_s.
    \label{eq:mixing}
\end{equation}
To properly reflect this mixture, the timestep conditioning for the student is applied heterogeneously per token---using $s$ where $M=1$ and $t$ where $M=0$---whereas the teacher is uniformly conditioned on $s$. 
Consequently, the student receives a random mosaic of cleaner and noisier tokens. 
To successfully match the teacher's robust representations, the student is forced to aggregate global context from cleaner spatial-temporal neighbors to infer the missing semantics of the heavily corrupted regions, directly driving the formation of strong internal representations.

\subsection{Representation Loss}
\label{sec:method_reploss}

Given the asymmetric inputs formulated in Sec.~\ref{sec:method_asymmetry}, we extract hidden states from two index sets of transformer blocks~\cite{dit,sd3} $\mathcal{S} = \{s_1, \dots, s_L\}$ (student) and $\mathcal{T} = \{t_1, \dots, t_L\}$ (teacher), each yielding tensors of shape $[B, N, D]$. 
We deliberately select teacher layers that are deeper than their corresponding student layers. 
As observed in recent self-supervised generative frameworks~\cite{sra,selfflow}, robust semantic representations naturally emerge in deeper network layers. 
Drawing targets from deeper teacher layers thus provides a stronger semantic distillation signal, while applying the loss to shallower student layers prevents the representation objective from interfering with the model's final generative outputs.

To bridge the representation spaces, a per-layer projector $\pi_\ell(\cdot)$ maps the student features into the teacher's domain:
\begin{equation}
    \tilde{h}^{\text{stu}}_{s_\ell} = \pi_\ell\bigl(h^{\text{stu}}_{s_\ell}\bigr),
\end{equation}
which is implemented as a multi-layer perceptron (MLP) with input/output widths equal to the backbone hidden size $D$ and an internal hidden width of $4D$.
We minimize the negative cosine similarity between projected student features and detached teacher features, averaged over tokens and layer pairs:
\begin{equation}
    \mathcal{L}_{\text{rep}}
    = \frac{1}{L} \sum_{\ell=1}^{L}
        \mathbb{E}_n\!\left[ 1 - \cos\bigl(\tilde{h}^{\text{stu}}_{s_\ell, n},\, \mathrm{sg}[h^{\text{tea}}_{t_\ell, n}]\bigr) \right],
    \label{eq:rep_loss}
\end{equation}
where $\mathrm{sg}[\cdot]$ denotes the stop-gradient operator. 
We opt for cosine similarity over $\ell_2$ distance because the EMA teacher and the student inherently operate at slightly different feature magnitude scales, making directional alignment more robust and appropriate.

\subsection{EMA Teacher and Projector Warmup}
\label{sec:method_ema}

\paragraph{EMA teacher.}
The teacher model $G_{\theta'}$ is initialized as a copy of the student and updated by the standard EMA rule
\begin{equation}
    \theta' \leftarrow \mu\, \theta' + (1 - \mu)\, \theta,
    \label{eq:ema}
\end{equation}
with the momentum $\mu$ set close to $1$ (e.g., $0.99$). 
Consequently, the teacher tracks a slowly moving, temporally smoothed version of the student.
Furthermore, deriving the teacher directly from the student's own parameter history ensures that the target representations remain strictly compatible with the model's native capacity, avoiding the distribution mismatch often encountered when distilling from disparate external models. 
Crucially, this alignment is entirely self-bootstrapped, eliminating the need for any external pre-trained vision encoders.

\paragraph{Projector warmup.}
Since distillation initializes both the student and the EMA teacher from a pre-trained backbone, whereas the projectors $\{\pi_\ell\}$ are initialized from scratch, the projectors inevitably output near-random vectors during the earliest training steps. 
Enabling the full objective at step~$0$ would undesirably backpropagate projector noise into the pretrained backbone and corrupt the delicate features we intend to align. 
To circumvent this, we implement a $T_{\text{proj}}$-step projector warmup phase. 
During this phase, $\mathcal{L}_{\text{DMD}}$ is temporarily disabled, and the student's hidden states $h^{\text{stu}}_{s_\ell}$ are detached before entering the projectors $\pi_\ell$. 
Thus, only the projectors receive gradients. 
The teacher is kept strictly identical to the student, ensuring both backbones remain stationary while the projectors reliably calibrate against fixed targets. 
After the warmup concludes at step $T_{\text{proj}}$, the full objective (Eq.~\ref{eq:total_loss}) is activated, and the EMA update (Eq.~\ref{eq:ema}) begins. 
At this boundary, we also synchronize the projector learning rate with that of the backbone, ensuring all representation-related parameters optimize at a unified pace once the student is unfrozen.

\section{Experiments}
\label{sec:experiments}

To evaluate the effectiveness of \NickNameShort{}, we structure our experiments to answer three core questions: (1) How do the asymmetric masking strategies influence representation learning across different generation paradigms? (Sec.~\ref{sec:exp_ablation}); (2) Does the representation alignment translate into perceptually coherent motions and detailed spatial structures? (Sec.~\ref{sec:exp_qualitative}); and (3) Does \NickNameShort{} quantitatively outperform recent state-of-the-art multi-step and distilled video generators? (Sec.~\ref{sec:exp_main}).

\subsection{Experimental Setup}
\label{sec:exp_setup}

\paragraph{Implementation details.}
We instantiate \NickNameShort{} across both bidirectional and causal paradigms to demonstrate its versatility.
For the bidirectional setting, we employ two structurally distinct backbones: Wan2.1-T2V-1.3B~\cite{wan} as our primary model, and HunyuanVideo-1.5~\cite{hunyuanvideo15} to evaluate generalizability.
Notably, HunyuanVideo-1.5 utilizes an MMDiT~\cite{sd3} architecture with strong text-rendering capabilities, serving as a stringent visual probe for spatial fidelity preservation during few-step distillation. 
For the causal autoregressive setting, we initialize the student from Causal-Forcing~\cite{causalforcing} ODE-regression checkpoint, enabling us to evaluate the chunk-by-chunk representation forcing under a genuine key-value (KV) cache rollout.

All configurations follow the established recipe~\cite{dmd2,selfforcing}, wherein the two score critics are drawn from the student's model family. 
Specifically, for the Wan2.1-based students (both bidirectional and causal), we use the frozen Wan2.1-14B~\cite{wan} as the real-score model and initialize the learnable fake-score model from Wan2.1-1.3B. 
For the HunyuanVideo-1.5 bidirectional student, the real-score model is the frozen HunyuanVideo-1.5~\cite{hunyuanvideo15} and the fake-score model is initialized from the same HunyuanVideo-1.5 checkpoint. 
We use $K{=}4$ denoising steps for both bidirectional and causal students.
To ensure structural consistency across different backbones, the \NickNameShort{} branch extracts features based on fixed depth ratios rather than absolute layer indices: the student and teacher features are read at depths $\ell_\theta{=}0.3 N_b$ and $\ell_{\theta'}{=}0.7 N_b$, respectively, where $N_b$ is the total number of transformer blocks. 
These features are aligned via an MLP projector (hidden dimension $4D$) with a representation loss weight $\lambda_{\text{rep}}{=}0.5$. 
The EMA teacher uses momentum $\mu{=}0.99$ with a $100$-step projector-only warmup, during which the projector's learning rate is amplified $50\times$. 
Optimization is performed using AdamW at $\text{lr}_G{=}2{\times}10^{-6}$ and $\text{lr}_C{=}4{\times}10^{-7}$ at a $5{:}1$ critic-to-generator update ratio.

\paragraph{Training cost.}
To quantify the overhead of \NickNameShort{} on top of vanilla DMD, we profile one training cycle (1 generator update + 5 critic updates, per the $5{:}1$ ratio). 
Table~\ref{tab:profile} reports the wall-clock breakdown by phase and component. The \NickNameShort{} branch --- EMA teacher forward, projector pass, and cosine alignment --- accounts for $\sim$$6\%$ of the cycle, while the remaining $\sim$$94\%$ is consumed by the DMD generator and critic update that the baseline already pays. 
\NickNameShort{} is therefore essentially free: the cost of the additional supervisory signal is dwarfed by the score-distillation loop that surrounds it.

\begin{table}
\centering
\caption{
\textbf{Per-cycle training-time breakdown (\% of one cycle).}
One cycle $=$ 1 generator update + 5 critic updates. The \NickNameShort branch is only $\sim$$6\%$ of the cycle.
}
\label{tab:profile}
\resizebox{\linewidth}{!}{%
\begin{tabular}{l cccc ccc c}
\toprule
 & \multicolumn{4}{c}{Generator update ($\times 1$)} & \multicolumn{3}{c}{Critic update ($\times 5$)} & \multirow{2}{*}{Total} \\
\cmidrule(lr){2-5} \cmidrule(lr){6-8}
Component & Rollout & DMD & \textbf{Ours} & Backward & Rollout & MSE & Backward & \\
\midrule
\% of cycle & 4.4 & 20.1 & \textbf{6.2} & 6.4 & 25.0 & 8.7 & 29.2 & 100.0 \\
\bottomrule
\end{tabular}}
\end{table}

To quantify the computational overhead introduced by \NickNameShort{}, we profile a complete training cycle (one generator update and five critic updates) built on top of vanilla DMD.

\begin{figure}
\centering
\includegraphics[width=\linewidth]{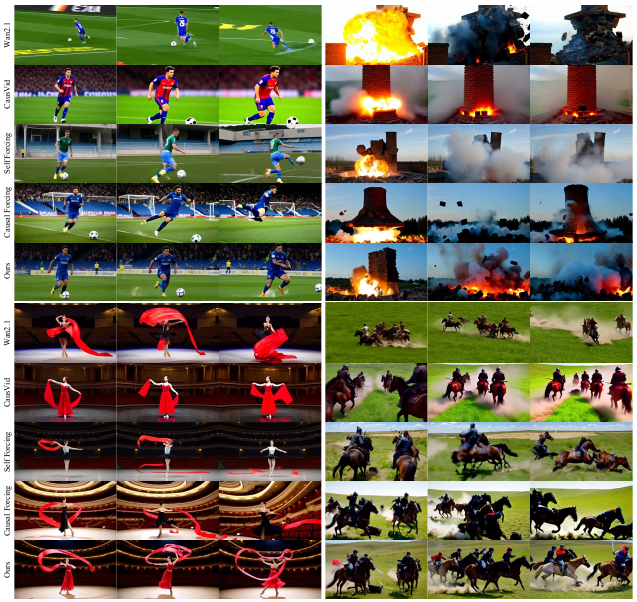}
\caption{
\textbf{Qualitative comparison on causal few-step T2V generation.}
Top row: multi-step bidirectional Wan2.1 teacher. Rows 2--5: few-step distilled causal students (CausVid, Self-Forcing, Causal-Forcing, \NickNameShort{}). 
The four panels expose four facets of representation-level coherence that the rep loss is designed to compress into the student: articulated-motion plausibility (top-left), event irreversibility (top-right), object persistence (bottom-left), and scene-structural stability (bottom-right). \NickNameShort{} satisfies all four, whereas baselines fail in distinct ways on each.
}
\label{fig:compare}
\end{figure}

\paragraph{Evaluation metrics.}
Our primary evaluation protocol is VBench~\cite{vbench}: we run the official prompt suite and report the Total Score, Quality Score, and Semantic scores. 
To complement these distributional metrics with a per-sample measure of human visual preference, we additionally report the VisionReward~\cite{visionreward} score --- a learned reward model trained to predict human preferences over generated videos, which is computed on a curated $100$-prompt set with rich motion and complex actions provided by~\cite{causalforcing}.
We further report Throughput (samples/s) and Latency (s/sample) on a single H100 GPU. 
For ablations we use VBench Quality as the headline single-number proxy.

\subsection{Ablation Experiments}
\label{sec:exp_ablation}

We study the two design choices introduced by the asymmetric input construction (Sec.~\ref{sec:method_asymmetry}): the masking ratio $\rho_{\max}$ and the mask sampling pattern. 
All curves below report VBench-style scores --- evaluated using the VBench~\cite{vbench} metric heads on a held-out validation set of $400$ motion-rich prompts generated by GPT-5 --- as a function of training step, which exposes both the final performance and the convergence behavior of each variant.

\begin{figure}
\centering
\includegraphics[width=\linewidth]{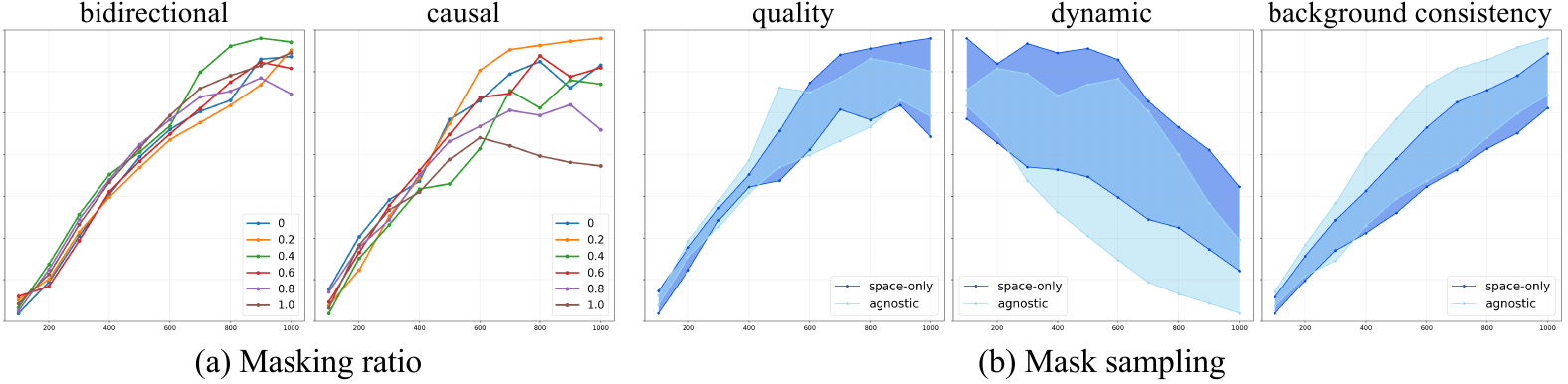}
\caption{
\textbf{Hyperparameter selection for asymmetric inputs.}
\textbf{(a) Masking ratio $\rho_{\max}$.} The two paradigms peak at different optima --- $\rho_{\max}{=}0.4$ for bidirectional and $\rho_{\max}{=}0.2$ for causal --- which we adopt as defaults in the following experiments.
\textbf{(b) Mask sampling pattern.} 
Space-only sampling (a single $H_p\times W_p$ mask broadcast across all frames) versus agnostic sampling (independent per-token masking) on VBench's Quality Score, Dynamic Degree, and Background Consistency. 
}
\label{fig:mask}
\end{figure}

\paragraph{Masking ratio.}
Fig.~\ref{fig:mask}(a) sweeps the the masking ratio $\rho_{\max} \in \{0, 0.2, 0.4, 0.6, 0.8, 1.0\}$ on both settings (left: bidirectional Wan2.1-1.3B; right: causal Wan2.1-1.3B initialized from Causal-Forcing). 
The two paradigms differ in where this optimum sits: the bidirectional student model peaks at $\rho_{\max}{=}0.4$, while the causal student model peaks at $\rho_{\max}{=}0.2$. 
The shift towards a smaller optimum in the causal setting reflects its narrower per-step token budget: each forward pass only sees the history chunks, so at the same $\rho_{\max}$ fewer clean tokens remain to anchor the alignment, and the rep target quickly becomes too noisy. 
The bidirectional student, which denoises the full window at once, has a much larger pool of clean tokens and tolerates a more aggressive mask.

\begin{table}
\centering
\caption{
\textbf{Quantitative comparison with existing methods on VBench~\cite{vbench} and VisionReward~\cite{visionreward}.} Models are grouped by sampling regime --- bidirectional multi-step, distilled bidirectional, autoregressive multi-step, and distilled autoregressive --- and \textbf{bold} marks the best score within each comparable group. 
Throughput is in FPS and latency in second, both measured on a single H100. 
All distilled models (in both distilled groups) share the same Wan2.1-T2V-1.3B backbone.
}
\label{tab:main}
\resizebox{\linewidth}{!}{%
\begin{tabular}{lcccccccc}
\toprule
Model & Params & Throughput $\uparrow$ & Latency $\downarrow$ & Resolution & Total $\uparrow$ & Quality\ $\uparrow$ & Semantic\ $\uparrow$  & Vision.\ $\uparrow$  \\
\midrule
\multicolumn{8}{l}{\emph{Bidirectional Video Diffusion Models}} \\
LTX-Video~\cite{ltxvideo}       & 1.9B  & 8.98 & 13.5 & 768$\times$512 &  79.83 & 81.88 & 71.62 & –6.218  \\
Wan2.1~\cite{wan}    & 1.3B & 0.78 & 103 & 848$\times$480 & 83.37 & 84.30 & 79.65 & 5.275   \\
\midrule
\multicolumn{8}{l}{\emph{Distilled Bidirectional Video Models}} \\
CausVid~\cite{causvid}    & 1.3B & 16.2 & 4.22 & 848$\times$480 & 83.19 & 84.08 & 79.63 & 5.310  \\
Ours & 1.3B & 16.2 & 4.22 & 848$\times$480 & \textbf{84.15} & \textbf{84.92} & 81.07 & \underline{7.060}  \\
\midrule
\multicolumn{8}{l}{\emph{Autoregressive Video Diffusion Models}} \\
Pyramid Flow~\cite{pyramidflow} & 2B & 6.70 & 2.5 & 640$\times$384 & 80.75 & 83.41 & 70.11 & 4.055  \\
SkyReels-V2~\cite{skyreelsv2} & 1.3B & 0.49 & 112 & 960$\times$540 & 81.97 & 83.96 & 74.01 & 3.584  \\
MAGI-1~\cite{magi1}      & 4.5B & 0.19 & 282 & 848$\times$480 & 78.88 & 81.67 & 67.72 & 0.773  \\
\midrule
\multicolumn{8}{l}{\emph{Distilled Autoregressive Video Models}} \\
CausVid~\cite{causvid}         & 1.3B  & \textbf{17.0} & \textbf{0.69} & 848$\times$480 & 81.33 & 83.98 & 70.72 & 5.741   \\
Self-Forcing~\cite{selfforcing} & 1.3B  & \textbf{17.0} & \textbf{0.69} & 848$\times$480 & 83.74 & 84.48  & 80.77 & 5.820  \\
Causal-Forcing~\cite{causalforcing} & 1.3B & \textbf{17.0} & \textbf{0.69} & 848$\times$480 & 84.04 & 84.59 & \underline{81.84} & 6.326\\
Ours & 1.3B & \textbf{17.0} & \textbf{0.69} & 848$\times$480 & \textbf{84.15} & \underline{84.67} & \textbf{82.06} & \textbf{7.523} \\
\bottomrule
\end{tabular}
}
\end{table}

\paragraph{Mask sampling pattern.}
Fig.~\ref{fig:mask}(b) compares the two mask sampling patterns~\cite{mae_st}: 
(i)~\textbf{space-only}, where a single $H_p\times W_p$ binary mask is sampled and broadcast across all frames in the window;
and (ii)~\textbf{agnostic}, where each spatio-temporal token is masked independently. 
To decouple the comparison from any specific choice of $\rho_{\max}$, we run each scheme on the causal student at four masking ratios $\rho_{\max}\in\{0.2, 0.4, 0.6, 0.8\}$, and plot the $[\min, \max]$ envelope across the four runs as a function of training step. 
A higher band therefore reflects an improvement that is consistent across the entire ratio range, not the artifact of a single lucky setting.

The two patterns reach comparable Quality (with space-only attaining a higher ceiling at convergence), but split on the two temporal-coherence metrics: 
space-only yields a higher Dynamic Degree, while agnostic produces a higher Background Consistency. 
We attribute this trade-off to a temporal-copy shortcut: under agnostic sampling, each masked token is likely left clean at the same spatial location in an adjacent frame, so the student can minimize the rep loss by copying along the time axis --- biasing it toward over-similar adjacent frames, which inflates background consistency and suppresses motion. 
Space-only sampling closes the shortcut: with the same mask broadcast across all frames, every spatial location is either clean or masked throughout the entire window, forcing reconstruction from spatial neighbors instead of from another time step. We adopt space-only as the default.

\subsection{Qualitative Results}
\label{sec:exp_qualitative}

\paragraph{Causal autoregressive.}
Fig.~\ref{fig:compare} compares autoregressive rollouts on identical prompts and seeds for the multi-step Wan2.1 teacher and four few-step distilled causal students: CausVid~\cite{causvid}, Self-Forcing~\cite{selfforcing}, Causal-Forcing~\cite{causalforcing}, and \NickNameShort{}. 
The improvement of \NickNameShort{} over the baselines is the global, physically-grounded structure that the rep loss compresses into the student's hidden states, which surfaces as four distinct facets of representation-level coherence across the figure. 
{(Top-left) Articulated-motion plausibility:} the soccer player's pose stays anatomically valid frame to frame, while baselines produce snapping or rerouted limbs. {(Top-right) Event irreversibility:} the destroyed wall stays destroyed, whereas baselines reconstitute or even regrow it across consecutive frames. {(Bottom-left) Object persistence:} the red ribbon remains a single continuous object throughout the dance, instead of vanishing, splitting, or warping. {(Bottom-right) Scene-structural stability:} horse and rider counts, identities, and overall layout stay coherent. Each baseline failure is a different symptom of the same missing internal abstraction, and \NickNameShort{} addresses all four through a single representation-level objective.

\begin{figure}
\centering
\includegraphics[width=\linewidth]{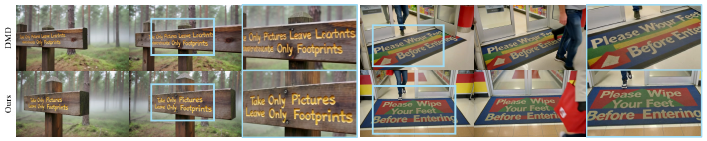}
\caption{
\textbf{Qualitative comparison on bidirectional few-step T2V generation.}
Both compared methods are distilled from the same HunyuanVideo-1.5 backbone. 
Each pair shows the DMD student (top) and \NickNameShort{} (bottom) on the same prompt and noise; 
a zoomed crop on the right of each panel highlights the rendered text. Across all four prompts, \NickNameShort{} produces sharp, legible glyphs where DMD collapses into blurred or scrambled characters, indicating that the rep loss strengthens the student's spatial representation.
}
\label{fig:compare_bid}
\end{figure}

\paragraph{Bidirectional T2V.}
Fig.~\ref{fig:compare_bid} compares two few-step HunyuanVideo-1.5 students --- the plain DMD baseline and \NickNameShort{} --- on prompts containing in-scene text. 
We use HunyuanVideo-1.5 here precisely because its multi-step teacher already renders non-trivial text; any degradation under distillation therefore directly probes how much fine spatial structure the student retains. 
The DMD student loses this ability: glyphs become blurred, distorted, or scrambled into wrong characters, indicating that output-only supervision is too coarse to preserve the highly structured spatial features that text requires. 
Adding the rep loss restores it: \NickNameShort{} produces sharp, legible text across all four prompts. 
Since text rendering is a stringent visual probe of spatial representation quality, this is direct evidence that \NickNameShort{} strengthens the student's internal spatial features.

\subsection{Quantitative Comparison}
\label{sec:exp_main}

Table~\ref{tab:main} reports the main comparison. 
At matched throughput and latency, \NickNameShort{} attains the best score on every reported metric within both the \emph{Distilled Autoregressive} and \emph{Distilled Bidirectional} groups.
We attribute these consistent gains to the rep loss directly stabilizing the student's intermediate features, which simultaneously raises per-frame fidelity and semantic / motion coherence without compromising any individual axis.

\begin{figure}
\centering
\includegraphics[width=\linewidth]{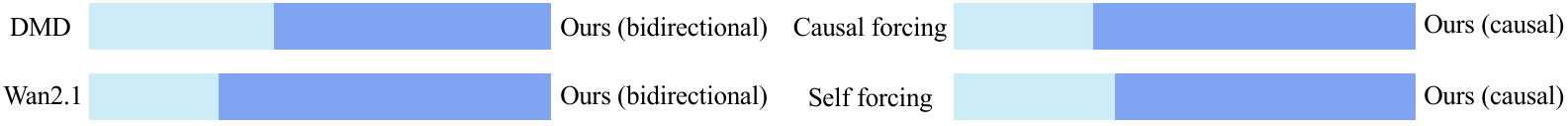}
\caption{
\textbf{Pairwise quantitative comparison.}
All compared methods share the same Wan2.1-T2V-1.3B backbone.
Each bar shows the per-prompt preference of VisionReward~\cite{visionreward}; the 
\textcolor[RGB]{127, 165, 242}{dark-blue}
segment is the fraction of prompts on which \NickNameShort{} is preferred. 
\NickNameShort{} wins the majority of prompts against all four baselines, on both bidirectional (DMD, multi-step Wan2.1 base model) and causal (Causal-Forcing, Self-Forcing) sides.
}
\label{fig:visionreward}
\end{figure}

\paragraph{Direct pairwise comparison.}
The aggregate VisionReward scores in Table~\ref{tab:main} are averaged over the prompt set; to confirm that the advantage of \NickNameShort{} is consistent at the prompt level rather than driven by a handful of outliers, we run a test on the curated $100$-prompt motion-rich set. 
For each prompt we generate two videos and let VisionReward~\cite{visionreward} pick the preferred one. 
Four pairings are evaluated, spanning both paradigms. As shown in Fig.~\ref{fig:visionreward}, \NickNameShort{} is preferred on the majority of prompts against every baseline, including the Wan2.1 base model.

\begin{figure}[h]
\centering
\includegraphics[width=\linewidth]{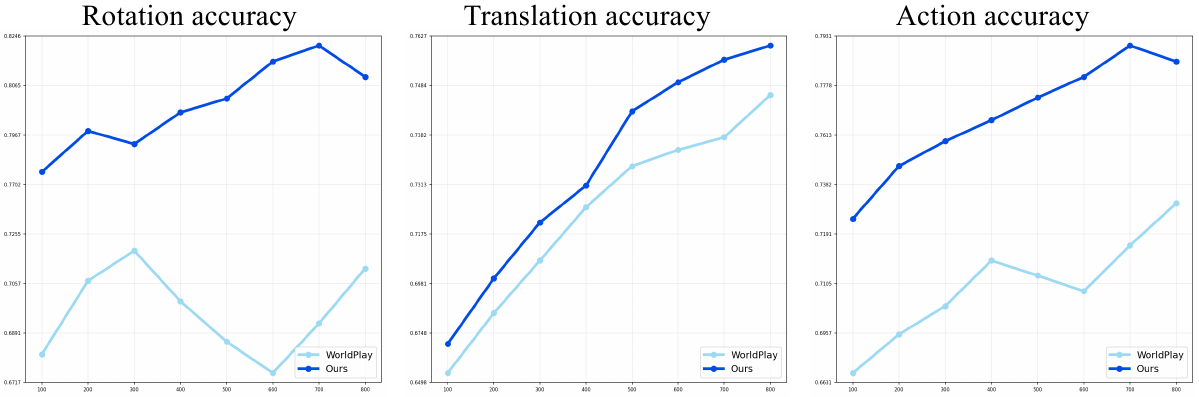}
\caption{
\textbf{Quantitative evaluation on interactive video generation.}
We evaluate the adherence of the generated video to the input camera and action conditions. Compared to the WorldPlay baseline, \NickNameShort{} achieves notably higher translation, rotation, and action accuracy across training steps, demonstrating that a stronger internal representation leads to more precise state transitions in interactive environments.
}
\label{fig:worldplay}
\end{figure}

\begin{figure}[h]
\centering
\includegraphics[width=\linewidth]{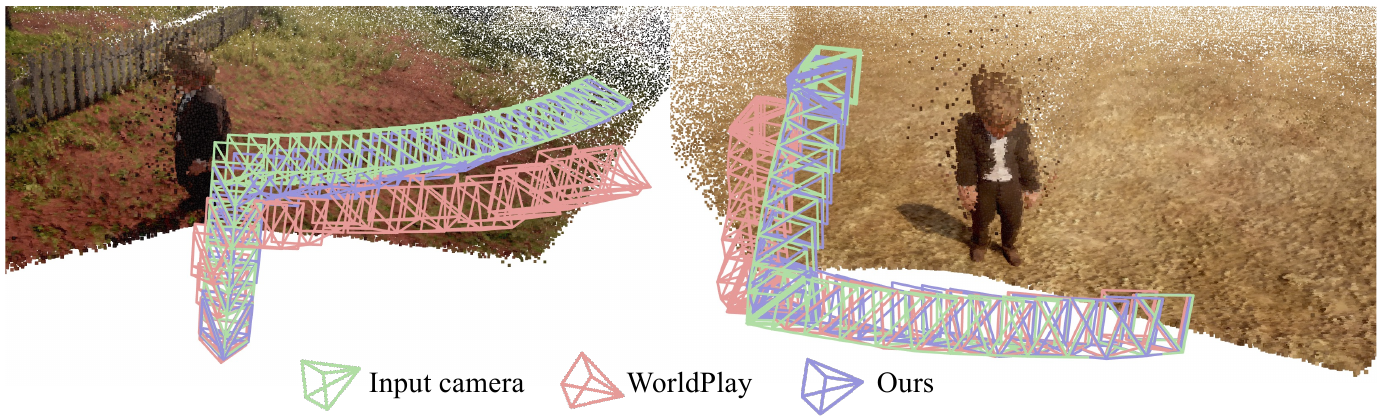}
\caption{
\textbf{Qualitative camera trajectory alignment.}
Reconstructed 3D point clouds and camera trajectories from the generated videos. The trajectory produced by \NickNameShort{} (\textcolor[RGB]{153, 148, 221}{blue}) tightly tracks the input condition (\textcolor[RGB]{177, 222, 164}{green}), whereas the WorldPlay baseline (\textcolor[RGB]{230, 155, 152}{red}) exhibits severe drift. This visualizes how the stronger internal representation of \NickNameShort{} enforces strict adherence to complex control signals.
}
\label{fig:worldplay_vis}
\end{figure}

\paragraph{Probing representation quality.}
We curate $120$ prompts depicting static scenes under smooth camera trajectories, reconstruct each generated video into a 3D Gaussian Splatting~\cite{3dgs} representation via WorldMirror-2~\cite{hyworld2}, re-render along the original trajectory, and report PSNR / SSIM / LPIPS against the generated frames, alongside VBench's Temporal Flickering and Subject Consistency as direct measures of frame-level coherence. 
On top of HunyuanVideo-1.5, \NickNameShort{} improves over vanilla DMD on every reconstruction and coherence metric at matched Dynamic Degree (Table~\ref{tab:3d}). 
Since no 3D supervision is used at any stage of training, this gain is attributable solely to the feature-level alignment of Sec.~\ref{sec:method}, providing direct evidence that the rep loss shapes the student's internal representation rather than only its output distribution.

\begin{table}[h]
\centering
\caption{
\textbf{Probing representation quality via re-rendering and frame-level coherence.} 
We report PSNR / SSIM / LPIPS against the generated frames, plus VBench Temporal Flickering and Background Consistency. \NickNameShort{} improves over vanilla DMD on every reconstruction and coherence metric at matched Dynamic Degree, despite using no 3D supervision during training.
}
\label{tab:3d}
\resizebox{\linewidth}{!}{
\begin{tabular}{lcccccccc}
\toprule
Model & Params & Throughput & PSNR $\uparrow$ & SSIM $\uparrow$ & LPIPS $\downarrow$ & Dynamic $\uparrow$ & Flickering $\uparrow$ & Consistency $\uparrow$ \\
\midrule
DMD (Ours w/o $\mathcal{L}_{\text{rep}}$)   & 8.3B & 2.21 & 20.64 & 0.6982 & 0.2144 & 90.00 & 96.39 & 94.11 \\
Ours  & 8.3B & 2.21 & \textbf{21.24} & \textbf{0.7110} & \textbf{0.2096} & 90.00 & \textbf{96.86} & \textbf{94.65} \\
\bottomrule
\end{tabular}
}
\end{table}

To rigorously quantify the robustness of the learned internal representations, we test how faithfully a generated video can be reconstructed as a unified 3D scene. 
Specifically, if the model has learned a strong internal representation, its generated videos will maintain strict temporal consistency, which naturally leads to highly accurate 3D reconstructions with minimal rendering error.
We curate $120$ prompts featuring static scenes navigated by smooth camera trajectories. 
Each generated video is lifted into a 3D Gaussian Splatting~\cite{3dgs} space via WorldMirror-2~\cite{hyworld2} and re-rendered along its original camera path. 
We report the reconstruction fidelity (PSNR, SSIM, and LPIPS) against the original generated frames, alongside VBench's Temporal Flickering and Subject Consistency metrics to directly assess frame-level continuity. 
As reported in Table~\ref{tab:3d}, \NickNameShort{} (applied to the HunyuanVideo-1.5 backbone) strictly surpasses the vanilla DMD baseline across every reconstruction and coherence metric, while maintaining an equivalent Dynamic Degree. 
Crucially, since no 3D supervision is injected at any stage of training, these geometric and temporal improvements are solely attributable to the feature-level alignment formulated in Sec.~\ref{sec:method}. 
This serves as compelling quantitative evidence that the \NickNameShort{} mechanism fundamentally enhances the student's intrinsic spatial-temporal awareness, moving beyond mere output-level distribution matching.

\paragraph{Application: Interactive long video generation.}
To further demonstrate the versatility of \NickNameShort{}, we extend our evaluation to an interactive video generation ~\cite{deepverse} setting using WorldPlay~\cite{worldplay}. 
WorldPlay is a causal autoregressive model that synthesizes interactive environments, conditioned on camera and action inputs. 
The original WorldPlay relies on an output-space distillation method akin to Self-Forcing~\cite{selfforcing}. 
We upgrade this distillation pipeline with \NickNameShort{} and evaluate the model's ability to follow complex control signals. 
For evaluation, we condition the models on a sequence of input camera trajectories, compute the actual synthesized camera poses from the generated long videos using WorldMirror-2~\cite{hyworld2}, and measure their alignment with the ground-truth input conditions. 
As illustrated in Fig.~\ref{fig:worldplay}, \NickNameShort{} achieves significantly higher translation accuracy, rotation accuracy, and overall action accuracy across training steps compared to the WorldPlay baseline. 
Furthermore, Fig.~\ref{fig:worldplay_vis} visualizes the the camera trajectories. The trajectory generated by \NickNameShort{} visibly conforms much closer to the input ground-truth camera path, whereas the baseline trajectory drifts. 
This affirms a core insight: interactive autoregressive generation is fundamentally a state-transition learning problem. 
By enforcing a more compact and structurally coherent internal representation, \NickNameShort{} equips the model to learn more precise state transitions, directly resulting in generated environments that strictly obey external conditional controls.

\section{Conclusion}
\label{sec:conclusion}

We presented \NickNameShort{}, which closes a structural blind spot of DMD by adding a predictive representation objective on top of its output loss. 
Through asymmetric inputs and a self-bootstrapped EMA teacher, the student is forced to align its internal features with the teacher's under information loss, supplying the representation-level signal DMD lacks. 
The mechanism is paradigm-agnostic and adds negligible cost, and consistently improves both spatial fidelity and temporal coherence across bidirectional and causal video generators on architecturally distinct backbones. We view \NickNameShort{} as evidence that, in the few-step regime, what the student learns to represent matters as much as what it learns to output.

\newpage
{
\small
\bibliographystyle{abbrvnat}
\bibliography{refs}
}

\newpage
\appendix
\section*{Appendix}


\section{Training Algorithm}
\label{sec:appendix_algorithm}

For completeness, we provide the full pseudo-code of one training iteration of \NickName{} in Algorithm~\ref{alg:repforcing}. The procedure mirrors the description in Sec.~\ref{sec:method}: each iteration reuses the standard DMD rollout to record a denoising trajectory, samples two trajectory states to construct the asymmetric student/teacher inputs, computes the joint loss $\mathcal{L}_{\text{total}} = \mathcal{L}_{\text{DMD}} + \lambda_{\text{rep}}\mathcal{L}_{\text{rep}}$, and updates the student, fake-score critic, projectors, and EMA teacher. The two warmup hyperparameters $T_{\text{proj}}$ and $T_{\text{ema}}$ control the projector-only and teacher direct-copy phases described in Sec.~\ref{sec:method_ema}.

\begin{algorithm}[t]
\caption{Training of \NickName.}
\label{alg:repforcing}
\begin{algorithmic}[1]
\Require Denoising timesteps $\{t_1, \dots, t_K\}$ with $t_1 > \dots > t_K$
\Require Student generator $G_\theta$, EMA teacher $G_{\theta'}$ ($\theta' \!\leftarrow\! \theta$)
\Require Fake-score critic $s_\theta$, frozen real-score $s_{target}$
\Require Per-layer projectors $\{\pi_\ell\}_{\ell=1}^{L}$, layer indices $\mathcal{S}, \mathcal{T}$
\Require Loss weight $\lambda_{\text{rep}}$, mask ratio $\rho_{\max}$, EMA momentum $\mu$
\Require Warmup steps $T_{\text{proj}}, T_{\text{ema}}$
\State $k \gets 0$
\Loop
    \State \textbf{// (1) DMD rollout with trajectory recording}
    \State Sample noise $z \sim \mathcal{N}(0, \mathbf{I})$ and condition $c$
    \State Roll out $G_\theta$ along $\{t_1, \dots, t_K\}$ and record
           $\mathcal{R} \!=\! \{(x_{t_i}, t_i)\}_{i=1}^{K}$,
           final prediction $\hat{x}_0$
    \State \textbf{// (2) Heterogeneous noise mixing}
    \State Sample two distinct steps $(t, s)$ from $\mathcal{R}$ with $t > s$
    \State Sample masking ratio $\rho \sim \mathcal{U}(0, \rho_{\max})$
    \State Sample per-token mask $M \in \{0,1\}^{B \times F \times H_p \times W_p}$,
           $M_{ijkl} \!\sim\! \mathrm{Bern}(\rho)$
    \State $\tilde{x}^{\text{stu}} \!\gets\! M \!\odot\! x_s + (1\!-\!M) \!\odot\! x_t$,\quad
           $\tilde{x}^{\text{tea}} \!\gets\! x_s$
    \State \textbf{// (3) Asymmetric forward and feature extraction}
    \State $\{h^{\text{stu}}_{s_\ell}\} \gets G_\theta(\tilde{x}^{\text{stu}}, t, c \,;\, \mathcal{S})$
           \Comment{with grad; chunk-AR if causal}
    \State $\{h^{\text{tea}}_{t_\ell}\} \gets G_{\theta'}(\tilde{x}^{\text{tea}}, s, c \,;\, \mathcal{T})$
           \Comment{no grad}
    \State \textbf{// (4) Loss computation}
    \If{$k < T_{\text{proj}}$}
        \Comment{Projector-only warmup: only $\{\pi_\ell\}$ receives gradient}
        \State $\mathcal{L}_{\text{DMD}} \gets 0$;\quad detach $h^{\text{stu}}_{s_\ell}$ before $\pi_\ell$ for all $\ell$
    \Else
        \State Compute $\mathcal{L}_{\text{DMD}}$ from $\hat{x}_0$ via score matching with $s_\theta, s_{target}$
    \EndIf
    \State $\mathcal{L}_{\text{rep}} \gets \dfrac{1}{L}\!\sum_{\ell=1}^{L}\! \mathbb{E}_n\!\left[1 - \cos\!\bigl(\pi_\ell(h^{\text{stu}}_{s_\ell, n}),\,\mathrm{sg}[h^{\text{tea}}_{t_\ell, n}]\bigr)\right]$
    \State $\mathcal{L}_{\text{total}} \gets \mathcal{L}_{\text{DMD}} + \lambda_{\text{rep}} \mathcal{L}_{\text{rep}}$
    \State \textbf{// (5) Parameter updates}
    \State Update $\theta$ and $\{\pi_\ell\}$ by backpropagating $\mathcal{L}_{\text{total}}$
    \State Update fake-score critic $s_\theta$ on $G_\theta$'s rollout (denoising loss)
    \If{$k < T_{\text{ema}}$}
        \State $\theta' \gets \theta$ \Comment{Direct copy: avoids fp32 EMA drift}
    \Else
        \State $\theta' \gets \mu\, \theta' + (1 - \mu)\, \theta$ \Comment{Standard EMA}
    \EndIf
    \State $k \gets k + 1$
\EndLoop
\end{algorithmic}
\end{algorithm}

\section{Application to Bidirectional and Causal Generators}
\label{sec:appendix_unification}

\NickNameShort{} is agnostic to the architectural form of the generator: the only assumption it makes about the backbone is that hidden states can be extracted at a configurable set of transformer blocks. Below we describe how the asymmetric inputs of Sec.~\ref{sec:method_asymmetry} are realized on each of the two paradigms studied in Sec.~\ref{sec:experiments}.

\paragraph{Bidirectional T2V.}
The entire latent window is denoised in a single forward pass. Following Sec.~\ref{sec:method_asymmetry}, we sample two trajectory states $(x_t, x_s)$ with $t > s$ from the DMD rollout $\mathcal{R} = \{(x_{t_k}, t_k)\}_{k=1}^{K}$ and form
\begin{equation}
    \tilde{x}^{\text{stu}} = M \odot x_s + (1 - M) \odot x_t, \qquad
    \tilde{x}^{\text{tea}} = x_s.
\end{equation}
The mixed input $\tilde{x}^{\text{stu}}$ is sent through the student $G_\theta$ at uniform timestep $t$, the cleaner input $\tilde{x}^{\text{tea}}$ through the EMA teacher $G_{\theta'}$ at $s$. The hidden states at the layer index sets $\mathcal{S}$ (student) and $\mathcal{T}$ (teacher) are read off in one shot and fed into the per-layer projectors $\{\pi_\ell\}$ for the cosine alignment of Eq.~\eqref{eq:rep_loss}. The added cost over standard DMD is exactly one extra teacher forward pass per iteration, independent of the number of denoising steps $K$.

\paragraph{Causal autoregressive generation.}
A causal student synthesizes the video chunk-by-chunk on its own previously generated context with a key--value (KV) cache. Naively running the rep loss over the full window would (i)~break the autoregressive structure that DMD is already trained against and (ii)~mix gradients across chunks that the student never sees jointly at inference. We instead replay the AR rollout at the representation level, mirroring the underlying Self-Forcing~\cite{selfforcing} pipeline:

\begin{enumerate}[leftmargin=1.4em, itemsep=2pt]
    \item \textbf{Per-pathway KV caches.} We maintain two independent KV caches, one for the student rollout and one for the teacher. Both are populated chunk-by-chunk and never share state, so the student gradient never touches teacher activations.
    \item \textbf{Chunk-local asymmetric forward.} For each chunk $b$, we feed the chunk-local mixed input $\tilde{x}^{\text{stu}}_b$ to the student (with grad) at chunk timestep $t$, and the chunk-local cleaner input $\tilde{x}^{\text{tea}}_b = x_s|_b$ to the teacher (no grad) at $s$. RoPE indices are kept chunk-aware so the positional encoding matches the AR rollout exactly.
    \item \textbf{Chunk-wise feature extraction.} The hidden states at $\mathcal{S}$ and $\mathcal{T}$ are collected per chunk and concatenated along the token axis only at the loss-computation step. The projector $\pi_\ell$ and cosine objective of Eq.~\eqref{eq:rep_loss} are then applied identically to the bidirectional case.
\end{enumerate}

\section{Hyperparameters and Implementation Details}
\label{sec:appendix_hparams}

Table~\ref{tab:appendix_hparams} reports the full set of hyperparameters for each of the three settings on which \NickNameShort{} is instantiated: the Wan2.1-T2V-1.3B bidirectional student, the HunyuanVideo-1.5 bidirectional student, and the Wan2.1-1.3B causal student initialized from a Causal-Forcing~\cite{causalforcing} ODE-regression checkpoint.

\begin{table}[h]
    \centering
    \caption{
    \textbf{Full hyperparameters for the three \NickNameShort{} instantiations.} The \NickNameShort{} branch is identical in spirit across settings; the only paradigm-dependent quantities are the masking ratio $\rho_{\max}$ and the chunk size $F_b$ (causal only). All bidirectional models use $K{=}4$ denoising steps; the causal student also uses $K{=}4$ unless otherwise noted.
    }
    \label{tab:appendix_hparams}
    \resizebox{\linewidth}{!}{%
    \begin{tabular}{lccc}
    \toprule
    & Wan2.1-1.3B (Bid.) & HunyuanVideo-1.5 (Bid.) & Wan2.1-1.3B (Causal) \\
    \midrule
    \multicolumn{4}{l}{\emph{Backbones}} \\
    Student backbone           & Wan2.1-T2V-1.3B   & HunyuanVideo-1.5  & Wan2.1-T2V-1.3B \\
    Real-score critic (frozen) & Wan2.1-T2V-14B    & HunyuanVideo-1.5  & Wan2.1-T2V-14B \\
    Fake-score critic (init.)  & Wan2.1-T2V-1.3B   & HunyuanVideo-1.5  & Wan2.1-T2V-1.3B \\
    Student init.\ ckpt        & Wan2.1-T2V-1.3B   & HunyuanVideo-1.5  & Causal-Forcing ODE-regression \\
    Backbone depth $N_b$       & $30$              & $54$              & $30$ \\
    \midrule
    \multicolumn{4}{l}{\emph{Diffusion / sampling}} \\
    Denoising steps $K$        & $4$               & $4$               & $4$ \\
    Timestep schedule          & flow-matching     & flow-matching     & flow-matching \\
    Timestep shift             & $7$               & $7$               & $7$ \\
    Guidance scale (training)  & $3.0$             & $3.0$             & $3.0$ \\
    Resolution                 & $848{\times}480$  & $848{\times}480$  & $848{\times}480$ \\
    Frames per video           & $81$ ($21$ latent) & $81$ ($21$ latent) & $81$ ($21$ latent) \\
    Chunk size $F_b$ (latent)  & --                & --                & $4$ \\
    \midrule
    \multicolumn{4}{l}{\emph{\NickName{}}} \\
    Layer pair $(\mathcal{S}, \mathcal{T})$  & $(\{9\}, \{21\})$ & $(\{18\}, \{42\})$ & $(\{9\}, \{21\})$ \\
    Depth ratio $(\ell_\theta, \ell_{\theta'})/N_b$ & $(0.30, 0.70)$ & $(0.33, 0.78)$ & $(0.30, 0.70)$ \\
    Projector $\pi_\ell$       & 3 layers MLP, hidden $4D$ & 3 layers MLP, hidden $4D$ & 3 layers MLP, hidden $4D$ \\
    Loss type                  & cosine ($1{-}\cos$) & cosine ($1{-}\cos$) & cosine ($1{-}\cos$) \\
    Rep weight $\lambda_{\text{rep}}$ & $0.5$        & $0.5$              & $0.5$ \\
    Mask ratio $\rho_{\max}$   & $0.4$              & $0.4$              & $0.2$ \\
    Mask sampling pattern      & space-only         & space-only         & space-only \\
    \midrule
    \multicolumn{4}{l}{\emph{EMA teacher and warmup}} \\
    EMA momentum $\mu$         & $0.99$            & $0.99$            & $0.99$ \\
    Projector-only warmup $T_{\text{proj}}$ & $100$ & $100$             & $100$ \\
    EMA direct-copy steps $T_{\text{ema}}$  & $100$ & $100$             & $100$ \\
    Projector lr multiplier (during warmup) & $50\times$ & $50\times$  & $50\times$ \\
    \midrule
    \multicolumn{4}{l}{\emph{Optimization}} \\
    Optimizer                  & AdamW             & AdamW             & AdamW \\
    $(\beta_1, \beta_2)$       & $(0.0, 0.999)$    & $(0.0, 0.999)$    & $(0.0, 0.999)$ \\
    Weight decay               & $0.01$            & $0.01$            & $0.01$ \\
    Generator lr               & $2{\times}10^{-6}$ & $2{\times}10^{-6}$ & $2{\times}10^{-6}$ \\
    Critic lr                  & $4{\times}10^{-7}$ & $4{\times}10^{-7}$ & $4{\times}10^{-7}$ \\
    Critic / generator updates & $5{:}1$           & $5{:}1$           & $5{:}1$ \\
    Mixed precision            & bf16              & bf16              & bf16 \\
    Grad checkpointing         & yes               & yes               & yes \\
    Max grad norm              & $10$              & $10$              & $10$ \\
    Per-device batch size      & $1$               & $1$               & $1$ \\
    Global batch size          & $8$               & $8$               & $8$ \\
    \midrule
    \multicolumn{4}{l}{\emph{Distributed training}} \\
    Sharding strategy          & FSDP (hybrid)     & FSDP (hybrid)     & FSDP (hybrid) \\
    Wrap policy                & size              & size              & size \\
    GPUs(global batch size)              & 32                & 32                & 32\\
    Training steps             & 750         & 750         & 750 \\
    \bottomrule
    \end{tabular}%
    }
\end{table}

\paragraph{Layer pair selection by depth ratio.}
We pick the layer pair $(\mathcal{S}, \mathcal{T})$ by fixed depth ratios rather than absolute indices~\cite{selfflow}, so the recipe transfers across backbones of different depth $N_b$. Concretely, we read the student feature at $\ell_\theta \approx 0.30\,N_b$ and the teacher feature at $\ell_{\theta'} \approx 0.70\,N_b$, which corresponds to $(\{9\}, \{21\})$ on Wan2.1's $30$-layer stack and $(\{18\}, \{42\})$ on HunyuanVideo-1.5's $54$-layer stack. The depth ratios are tuned once on Wan2.1 and applied to HunyuanVideo-1.5 without further search.

\paragraph{Projector design.}
Each per-layer projector $\pi_\ell$ has input/output widths equal to the backbone hidden size $D$ (no dimensional change between student and teacher) and an internal width of $4D$. We adopt the 3-layer MLP as default ~\cite{repa}.

\paragraph{Implementation details.}
We highlight a few implementation details that mattered for stability and that are easy to overlook:
\begin{itemize}[leftmargin=1.4em, itemsep=2pt]
    \item \textbf{Stop-gradient on the teacher path.} The teacher hidden states are detached before being fed into the cosine loss; without this, gradient leaking through $G_{\theta'}$ destabilizes the EMA update.
    \item \textbf{Projector lr reset at $T_{\text{proj}}$.} During warmup the projector lr is scaled by $50\times$ to let it catch up from random initialization. At the end of warmup we reset its lr multiplier to $1.0$ so all rep-related parameters are updated at a unified pace once the student is unfrozen. Failing to reset led the projector lr to dominate the joint phase and slowly diverge.
\end{itemize}

\section{Evaluation Protocol Details}
\label{sec:appendix_evaluation}

\paragraph{VBench~\cite{vbench}.}
We follow the official VBench T2V evaluation protocol. Each model generates $5$ videos per prompt over the full VBench prompt suite ($946$ prompts), and we report the official Total / Quality / Semantic scores together with the Dynamic Degree, Temporal Flickering, and Background Consistency dimensions wherever these are quoted in the main paper. Throughout we use the official VBench metric heads with no modification.

\paragraph{VisionReward~\cite{visionreward}.}
For the aggregate VisionReward score reported in Table~\ref{tab:main}, we use the curated $100$-prompt motion-rich set provided by Causal-Forcing~\cite{causalforcing}, generate one video per prompt at the model's standard resolution, and compute the average VisionReward score across all $100$ prompts. We report the score directly without any post-processing.
For Fig.~\ref{fig:visionreward}, we take the same $100$-prompt set and, for each prompt, generate two videos --- one with \NickNameShort{} and one with each baseline (DMD, multi-step Wan2.1, Causal-Forcing, Self-Forcing). We then ask VisionReward to score both videos and record the per-prompt preference (the higher-scoring video wins). Each bar in Fig.~\ref{fig:visionreward} reports the fraction of prompts on which \NickNameShort{} is preferred over the corresponding baseline; the prompt set, seed, and generation hyperparameters are fixed across all four pairings.

\paragraph{Validation set for ablations.}
The 400-prompt motion-rich validation set used in the ablation curves of Sec.~\ref{sec:exp_ablation} is generated by GPT-5 with explicit instructions to elicit prompts containing complex motion (articulated subjects, fast camera or subject movement, multi-object interaction, irreversible events). We use this set rather than the full VBench suite to (i)~stress motion --- the failure mode \NickNameShort{} is designed to address --- and (ii)~speed up the per-checkpoint evaluation, which we run every $100$ training steps to expose convergence behavior. The 400 prompts are held out from any training corpus.

\paragraph{3D re-rendering probe (Table~\ref{tab:3d}).}
For the representation-quality probe of Table~\ref{tab:3d}, we curate $120$ prompts depicting predominantly static scenes under smooth camera trajectories. 
For each prompt, the trained student generates a video, which is then reconstructed into a 3D Gaussian Splatting representation via WorldMirror-2~\cite{hyworld2}; the resulting 3DGS scene is re-rendered along the original camera trajectory. PSNR / SSIM / LPIPS are computed between the re-rendered and the original generated frames, and Temporal Flickering / Background Consistency are computed via VBench on the original generated frames. No 3D supervision is used at any stage of training, so any improvement in PSNR/SSIM/LPIPS is attributable to the representation-level objective rather than to direct 3D regularization. We additionally pin the Dynamic Degree of the two compared methods (the DMD baseline and \NickNameShort{}) so that any reduction in the residual cannot be ``bought'' by suppressing motion.

\section{Datasets}
\label{sec:appendix_datasets}

\paragraph{Training corpus.}
For all bidirectional and causal T2V training we use VidProM dataset~\cite{vidprom}. 

\paragraph{Evaluation prompt sets.}
Three prompt sets are used in the experiments:
\begin{itemize}[leftmargin=1.4em, itemsep=2pt]
    \item \textbf{VBench~\cite{vbench}}: the full $946$-prompt T2V suite, used for Table~\ref{tab:main} and any VBench-derived metric.
    \item \textbf{$100$-prompt motion-rich set} (from Causal-Forcing~\cite{causalforcing}): used for the aggregate VisionReward score in Table~\ref{tab:main} and the pairwise comparison in Fig.~\ref{fig:visionreward}.
    \item \textbf{$400$-prompt motion-rich validation set}: a held-out set generated by GPT-5 and used for all ablation curves.
    \item \textbf{$120$-prompt static-scene set}: a held-out set generated by GPT-5 and used for the 3D re-rendering probe in Table~\ref{tab:3d}.
\end{itemize}
None of these prompt sets are used during training.

\section{Limitations}
\label{sec:appendix_limitations}

\paragraph{Inherited capabilities and biases.}
\NickNameShort{} is a distillation-time objective: it does not add new capabilities to the student beyond what its multi-step teacher can express. Concretely, prompt adherence on long compositional prompts and on rare visual concepts is upper-bounded by the teacher's behavior, and any biases of the underlying training corpus (cultural skew, demographic representation, domain coverage) are inherited unchanged. The rep loss can sharpen the student's internal features but cannot manufacture concepts the teacher does not represent.

\paragraph{Backbone coverage.}
We instantiate \NickNameShort{} on two structurally distinct backbones (Wan2.1 and HunyuanVideo-1.5) and on both bidirectional and causal paradigms, but other architectural families --- e.g.\ purely-attention-free state-space backbones or hybrid architectures --- are out of scope of this paper and may require revisiting the layer-pair selection rule.

\end{document}